\documentstyle[aclap]{article}
\newcommand{\scare}[1]{`#1'}

\newcommand{\term}[1]{{\sc #1}}
\newcommand{\lingform}[1]{{\it #1}}
\newcommand{\italics}[1]{{\it #1}}

\author{Mark Dras \& Mike Johnson\\
Microsoft Research Institute \& Department of Computing\\
Macquarie University \\
North Ryde NSW 2109 \\
Australia \\{\tt \{markd,mike\}@mpce.mq.edu.au}
}

\title{Death and Lightness: \\
Using a Demographic Model to Find Support Verbs}

\begin{document}

\maketitle

\begin{abstract}
Some verbs have a particular kind of binary ambiguity: they can carry their normal,
full meaning, or they can be merely acting as a prop for the nominal object.  It has
been suggested that there is a detectable pattern in the relationship between a verb
acting as a prop (a \term{support verb}) and the noun it supports.

The task this paper undertakes is to develop a model which identifies the support verb for a particular noun, and by extension, when nouns are enumerated, a model which disambiguates a verb with respect to its support status.  The paper sets up a basic model as a standard for comparison; it then proposes a more complex model, and gives some results to support the model's validity, comparing it with other similar
approaches.
\end{abstract}

\section{Introduction}

It is well-known that some verbs have a binary ambiguity: consider the sentences \lingform{Kim took a photograph of Dale} and \lingform{Kim took a painting of Dale}.  The former can be a paraphrase of  \lingform{Kim photographed Dale} while the latter has no such paraphrase.  This is because, in one reading of the first example, the lexeme \lingform{take} is only acting as a prop or support for a content-bearing noun, a capacity first noted by Jespersen (1942); while in the second example the verb has only its full meaning of \lingform{gain possession}.  It has been noted that many support verbs (SVs), like \lingform{take} and \lingform{make}---as in \lingform{make a distinction} (equivalent to \lingform{distinguish})---are quite productive, being able to act as support verbs for a number of different nouns; and investigations have suggested that there may be a pattern in the relationship between SV and noun (e.g.  Makkai, 1977; Wierzbicka, 1982).  Discovering the correspond!
 ence between SV and noun can help disambiguate the verb, deciding whether it is acting as an SV or not; that is the aim of this paper.

The concept of ambiguity here is similar to that of word-level sense ambiguity (see, for example, Yarowsky, 1992), rather than to higher-level ambiguity such as that of garden-path sentences (Gibson, 1995).  Because an SV by itself does not represent a separate concept---the whole construction \lingform{take a walk} represents a single concept of walking---while a full verb can, removing this type of ambiguity from input is particularly important for determining mappings in areas where input text is translated into another form, such as to text in another language, in machine translation (Danlos and Samvelian, 1992), or to a meaning representation (Meteer, 1991); it is also useful for dealing with multiword constructions in language, like
idioms (Abeille, 1988; Storrer and Schwall, 1993).

More generally, identifying a verb as an SV indicates its lack of propositional
content, and 
so can contribute to more accurate readability measures, such as lexical 
density (described in Halliday, 1985).  Knowing whether a word lacks
content is similarly important in the area of information retrieval, in explicitly 
constructing stoplists (Salton, 1988), on the assumption that content-free
words should be deleted from the search space of key terms.  These lists
can be made
more comprehensive by recognising the similar lack of content in SVs and
the closed-class words, which are traditionally considered to comprise the
set of content-free words (Halliday, 1985).  Another area of potential use
is in style checking, where it is generally recommended that SVs be removed for reasons of clarity (Kane, 1983); for example, [1a] becomes [1b].

\begin{itemize}
\item[1a.] It is important for teachers to have a knowledge of their students.
\item[1b.] It is important for teachers to know their students.
\end{itemize}

Characterisation and identification techniques for SVs have used both
purely semantic methods and more syntactic, surface-based ones. An 
example of the former is given in
Wierzbicka's 1982 paper, where a set of semantic rules is proposed to 
determine the SV that corresponds to a particular noun, concentrating on 
explanations of phenomena like why one can \lingform{have a drink} but not 
\lingform{*have an eat}.  Her analysis of these phenomena leads to rules like:

\begin{itemize}
\item[2.] The support verb is \lingform{have} if the nominalisation represents an
action aiming at a perception which could
cause one to know something and which would not cause one to feel bad if it
didn't.
\end{itemize}

The surface-based approaches aim to overcome
the laborious nature of determining such semantic rules by assuming that the 
syntactic structure reflects enough of the semantics to make a surface 
statistical analysis possible.  Fontenelle (1993) proposed a surface-based
approach, which uses the work of Smadja (1991) on collocation
relations in text.  His method, however, requires multi-lingual machine readable
dictionaries, which may not be readily available, and the prior division of words into
sets according to Mel'cuk's Meaning-Text Theory (c.f. Mel'cuk and Zholkovsky,
1988; Steele, 1990).

A more recent example of the surface-based approach is the statistical 
technique proposed by Grefenstette and Teufel~(1995).  A statistical
analysis sounds intuitively plausible, 
given that it has been suggested (Halliday, 1985) that there is a relationship
between frequency of a word, a surface phenomenon, and 
its content-freeness.  However, Grefenstette and Teufel's (1995) statistical
technique only uses frequency with respect to a particular noun (which I call
\term{local information}), rather than any more general notion of frequency.
So, for example, in identifying an SV for \lingform{demand}, Grefenstette
and Teufel look only at the frequency of co-occurrence of various verbs
with \lingform{demand}.  As a result, in their system
\lingform{meet} is chosen as the corresponding SV
as it is the most frequently co-occurring verb; \lingform{make}, the actual
SV, is ranked lower.  However, knowing that \lingform{make} is a generally
productive SV would lead it to be a more obvious candidate, despite its lower
ranking.  
A probabilistic argument to this effect is given in section 3; the key approximation on which it relies, the use of data with respect to all other SV constructions
(\term{local information}), is drawn from a model in demographic statistics,
and is outlined first in section 2.

\section{Mortality}

This section looks at a standardisation model used in mortality which is a
useful one for SVs: it provides a way of combining information about a
subpopulation---the \term{target population}---with a larger population
which provides more information---the \term{standard population}. An
overview of the theory is given below;
more detail can be found in standard demographic texts such as
Pollard \italics{et al}~(1981).  The
method described in this paper combines local and global information
about SVs in a similar way; the correspondence will be discussed in
more detail in Section 3.

\subsection{Standard populations}

The most easily obtainable mortality rate, the crude death rate (CDR), is
calculated by dividing the total number of deaths for a population by the
total size of the population.  However, this does not accurately reflect the
mortality experienced by the population: Pollard \italics{et al}~(1981) discuss the
situation of Maori and non-Maori populations of New Zealand in 1966, where
the Maori population had a lower CDR despite having higher mortality rates
for every age group.

The explanation for this discrepancy comes from the different
profiles of each population: the age categories which experience the lowest
rates have higher population sizes, weighting the overall population rate
so that it also is lower.  So the Maori population has a much higher number of 
young people, who have lower death rates; this produces a lower overall rate, 
as the CDR effectively weights the measurement by the distribution of the 
Maori population.  Using a common (or standard) population is one way of
removing this bias.

\subsection{Indirect Standardised Death Rate (ISDR)}

One standard demographic technique for producing a figure comparable
between populations is to apply age-specific mortality rates from 
the standard population to the corresponding age brackets of the 
target population, giving the number of deaths that would be expected in the target 
population if the levels of mortality in the standard population were being
experienced.
These expected deaths are summed, and used in the calculation of the 
standardised mortality ratio, which is equal to actual deaths for the 
target population divided by expected deaths; it represents the degree above 
or below expectation to which deaths actually occurred.
This ISDR is often the preferred measure of standardisation when the target 
population is too small to accurately calculate age-specific mortality rates, 
using as it does those of the standard population in their place.

There is no one definitive standard population for two given target 
populations. One frequently chosen standard population is the 
union of the two target populations: for example, when comparing Maori 
mortality with non-Maori mortality in New Zealand, the total New Zealand 
population was used as standard.

\section{A Probabilistic Model}

This section describes a probabilistic model for the prediction of support
verbs, along with the approximations and assumptions being made; these
are then justified by recourse to the demographic model described in
section 2.

The most likely support verb for a given nominalisation is defined as 
that verb which has the highest probability of being an SV for that
nominalisation; taking the point estimate of this probability, the 
most likely support verb for a nominalisation is
that verb which has the highest frequency of occurrence as an SV with
the nominalisation.  That is:

\begin{eqnarray}
\mbox{\sc SV}(j) &=& \mbox{\it argmax}_{i \in V}f_{ij}
\end{eqnarray}

where

\begin{itemize}
\item $SV(j)$ = most likely SV for nominalisation $j$

\item $f_{ij}$ = frequency with which verb $i$ appears to be supporting nominalisation $j$ 
\end{itemize}

This quantity $f_{ij}$ is, of course, unknown, as there are no corpora tagged for verb
lightness---that tagging is the purpose of the identification method proposed in this paper and others.  Now, $f_{ij}$ can be rewritten as

\begin{eqnarray}
\mbox f_{ij} &=& \mbox m_{ij} p_{ij}
\end{eqnarray}

where

\begin{itemize}
\item $m_{ij}$ = number of occurrences of verb $i$ governing\footnote{
I use \scare{govern} in the sense that if X is a complement of Y, then Y governs X;
this is in line with Mel'cuk (1988).
}
nominalisation $j$

\item $p_{ij}$ = Pr (verb $i$ is acting as an SV $|$ nominalisation $j$)
\end{itemize}

\subsection{Basic model}

Again, $p_{ij}$ is unknown and cannot be estimated directly.  One approach is to
make the admittedly inaccurate assumption that $p_{ij}$ is equal to 1 for 
all $i$ and $j$.  That
is, the verb chosen to be the SV is simply that one which most frequently governs
the verb in the chosen training corpus.  Then

\begin{eqnarray}
\mbox f'_{ij} &=& \mbox m_{ij}
\end{eqnarray}

which gives

\begin{eqnarray}
\mbox{\sc SV}(j) &=& \mbox{\it argmax}_{i \in V}m_{ij}
\end{eqnarray}

Grefenstette and Teufel (1995) effectively use this assumption, with the additional modification of restricting the count $m_{ij}$, by only
counting those occurrences where the SV construction has similar characteristics
to the equivalent full verb.  So, for example, the preposition qualifying the noun
in the SV construction is generally the same as the preposition attached to the
full verb (\lingform{make a decision to ...}, \lingform{decide to ...}); they use
this type of information, when collecting data, to give a more accurate $m_{ij}$.

In this paper I will only be looking at the gain that can be made from attempting to
estimate $p_{ij}$, so I will be using this definition of $SV'(j)$
as the main basis for comparison.  I will, however, also compare
the results against the model of Grefenstette and Teufel (1995), to compare the
degree of improvement expected of each over the basic model.

\subsection{Global Information Model}

Now, an approximation for $p_{ij}$ suggested by the demographic model above
is to use the unconditional probability over
all nominalisations---call this $p_i$.  So:

\begin{eqnarray}
\mbox f''_{ij} &=& \mbox m_{ij} p_{i}
\end{eqnarray}

where

\begin{eqnarray}
\mbox p_i &=& Pr (verb.i.is.acting.as.an.SV) \nonumber \\
	&=&  \sum_j m_{ij} p_{ij} / \sum_j m_{ij}
\end{eqnarray}

The case for using such a significant approximation here is the same as the
case for using it in the demographic model
described in section 2; it is the approximation around which the model is
built.  The use of unconditional probabilities as
an approximation to the conditional ones parallels, in the demographic
model, the use of the standard or global population rates when calculating
statistics on the sub-or target population.  The correspondence between
elements of the two models is given in Table 1.

{\small
\begin{table*}
\begin{tabular} {|p{1.3in}|p{1.3in}||p{1.3in}|p{1.3in}|}
\hline
\multicolumn{2}{|c|}{Mortality} & \multicolumn{2}{|c|}{Lightness} \\
\hline
description & instance & description & instance \\
\hline
\hline
target population & Maori population & local information (data for
given nominalisation, such as \lingform{make}) & all verbs 
governing given nominalisation, such as \lingform{make} \\
\hline
standard population & NZ population & global information (data for
all nominalisations) & all verbs governing all nominalisations \\
\hline
age category & ages 15-25 & verb & instances of \lingform{make}
as a governing verb \\
\hline
target population mortality rate & & conditional probability $p_{ij}$ & \\
\hline
standard population mortality rate & & unconditional probability $p_i$ & \\
\hline
\end{tabular}
\caption{Correspondence between mortality and lightness models}
\label{tab2}
\end{table*}
}

\vspace{0.15in}

The reason behind using the global rates in both cases is similar---the
local probabilities cannot accurately be estimated.  A difference, however, 
can be noted.  In the demographic model, the local and global
probabilities of dying will both usually follow typical mortality curves:
high rates at birth, declining until late teens, an \scare{accident hump}
of higher mortality, another decline, and then increasing with middle
and old age.  But, in the language model, the conditional and unconditional
probabilities are less innately similar.  The conditional probabilities will generally
be more dichotomous: for a given nominalisation, a verb will either
(virtually) always or (virtually) never be an SV.  The global, unconditional 
probabilities will, on the other hand, be a more mixed distribution: for
example, \lingform{make} may have an unconditional probability of being
an SV of 0.3, \lingform{have}, a probability of 0.23, and so on.

It should also be noted, however, the mortality curves of the demographic
model can actually be very
dissimilar also---the mortality rate distribution for the target population may
lack an accident hump, or the probability of dying may approach 1 much faster
and earlier than in the global mortality distribution, producing an effect similar to
that in the language situation described.  The approximation technique is
fairly robust, to allow for this, and is not greatly affected by the choice of the
global population or rates (see Pollard, 1981: 72).  In any case, it is more
accurate than the assumption of
$p_{ij}$ equaling 1: the $p_i$ are a ranking of the likelihood of a verb being an SV when no context is known, meaning that more likely candidates can be identified,
whereas the basic model gives no such indication.

Estimating these unconditional probabilities relies on an assumption that
support verbs are productive to some extent, an assumption which appears to hold
true for at least the major support verbs---for example, \lingform{make} acts as
an SV for \lingform{attempt}, \lingform{criticism}, \lingform{decision},
\lingform{error}, \lingform{judgment}, and many other nominalisations---with
a corollary to the 
assumption, that other non-support verbs will not exhibit the same generality across
nominalisations, and this seems to be borne out by inspection of the global 
information described in section 4.

Then, given this assumption, the unconditional probabilities can be estimated by,
for the purposes of this estimation only, treating all occurrences of verbs
governing nominalisations in the corpus as acting as support verbs, and
aggregating these to give the unconditional probability.  That is:

\begin{eqnarray}
\mbox p'_i &=& \sum_j m_{ij} / \sum_i \sum_j m_{ij}
\end{eqnarray}

This can be thought of as producing a global ordering of verbs, in which the rankings approximate the likelihood of acting as an SV because of the productivity of support
verbs, which will leading to $p_i$ and $p'_i$
correlating reasonably well.  Productive support verbs will tend to govern a range of
different nominalisations and be ranked high in this ordering; their higher values
of $p'_i$ correspond to their higher likelihood of being support verbs as
measured by $p_i$.  A less productive SV, such as \lingform{bear}, will
still have a low probability estimate $p'_i$.  This approximation is not
expected to give accurate estimates for $p_i$; it is only important that it
correlate with $p_i$, as the process of choosing the most likely SV only
requires that the ranking of verbs in order of the probability of being an
SV be accurate.  Again, an inspection of the global information described
in section 4 seems to bear this out.

Given these approximations, the most likely SV under this model is given by:

\begin{eqnarray}
\mbox{\sc SV''}(j) &=& \mbox{\it argmax}_{i \in V}m_{ij} p_i \nonumber \\
&=& \mbox{\it argmax}_{i \in V}m_{ij} \sum_j m_{ij} / \sum_i \sum_j m_{ij} \nonumber \\
&=& \mbox{\it argmax}_{i \in V}m_{ij} \sum_j m_{ij}
\end{eqnarray}

\section{Experimental work}

\subsection{Deriving local and global information}

To gather both local and global information, the 1992 version of Grolier's 
encyclopedia of approximately 8 million words was used, tagged by the 
part-of-speech tagger developed by Brill~(1993).  A heuristic for producing the 
local information about a target population involved
searching the corpus for the nominalisation, determining the verb for which
the nominalisation was the direct object, and tallying the relative frequency
of these verbs.

Grefenstette and Teufel~(1995) note that a confounding factor in the local 
information, when picking out nominalisations and their governing verbs, is that 
the nominal may have become \term{concretised}. 
Generally, nominalisations represent an abstract concept, being essentially events
represented in noun form; but it is possible for the nominal to represent
a physical embodiment of that concept.
For example:

\begin{itemize}
\item [3a.] {\sc abstract:} He made his formal proposal to the full committee.
\item [3b.] {\sc concretised:} He put the proposal in the drawer.
\end{itemize}

The abstract and concretised versions will tend to have different governing 
verbs.  However, if the assumption about productivity in Section 3.2 is true, 
and the global information is a good approximation to the innate lightness 
of a verb, the correct SV will be favoured over those associated with the 
concretised forms.

\subsection{Generating nominalisations for global information}

To construct the global information, data for all nominalisations is needed.  
A large list of nominalisations was derived in a partially
automated manner from Longman's 
Dictionary of Contemporary English (LDOCE) using both built-in 
information and a heuristic: since a nominal is an event represented in noun 
form, 
the procedure used here for deriving a list of them involved looking 
for nouns with associated \term{stem verbs}; e.g., \lingform{decide} is 
the stem verb of \lingform{decision}.
Some verbs have this information encoded in their entries: for example, 
\lingform{adjust} lists \lingform{adjustment} as its nominalisation; 
there were 257 verbs in this category.  For others, 
an automatic orthographic heuristic that matched nouns with verbs produced a 
set of candidates, which was 
manually filtered to produce 1414 more deverbal nominalisations.

A set of support verb constructions and their constituent nominalisations
was drawn from a range of sources---see Table~\ref{tab3} and bibliography 
for references---and
used as the test set for the experiment.  \footnote{
These sources have assumed that the propositional meanings of the SV
construction and the full verb are equivalent.  This may be disputed in a
number of cases, but for the purposes of this paper, the equivalence of the
two meanings will be taken as indicated by the relevant source.
}
The list of nominals did not cover some of
the nominals from the test,
so the local information was generated from the training corpus for 
each of the missing test set nominals and aggregated into the global 
information.

\subsection{The test set and results}

A system to identify support verbs for nominalisations, based on the global
information model, was implemented
by tabulating the lemmatised forms of all the verbs for which these nominals 
were the direct object.
Candidate support verbs were ranked in order of values of SV''(j), and 
the maximum of these values chosen.

{\small
\begin{table*}
\begin{tabular} {|l|l|c|l|l|l|} 
\hline
Source Text & Verb & Choice C1 & Choice C2 & Ratio (C1/C2) & Reference \\
\hline
\hline
make an attempt & attempt & make & include & 9.36 & Dras, Dale (1995) \\
\hline
make a change & change & make & produce & 1.85 & Dras, Dale (1995)  \\
\hline
make a concession & concede & make & include & 11.47 & Dras, Dale (1995) \\
\hline
make a demand & demand & make & create & 1.03 & Gref., Teufel (1995) \\
\hline
make a distinction & distinguish & make & have & 3.04 & Meteer (1991) \\
\hline
have a drink (of) & drink & become & N/A & N/A & Wierzbicka (1982) \\
\hline
have an effect (on) & affect & have & produce & 3.04 & Dras, Dale (1995) \\
\hline
have a feeling & feel & have & produce & 3.27 & Harris (1957) \\
\hline
make a gift (of) & give & have & include & 9.89 & Harris (1957) \\
\hline
do harm (to) & harm & cause & do & 1.26 & Huddleston (1968) \\
\hline
make a judgment & judge & make & have & 2.43 & Dras, Dale (1995) \\
\hline
have a knowledge (of) & know & have & use & 12.36 & Kane (1983) \\
\hline
make progress & progress & make & allow & 64.33 & Harris (1957) \\
\hline
make a proposal & propose & make & include & 1.10 & Gref., Teufel (1995) \\
\hline
bear a resemblance (to) & resemble & bear & have & 2.64 & Huddleston (1968) \\
\hline
give a shove (to) & shove & N/A & N/A & N/A & Harris (1957) \\
\hline
have a snooze & snooze & N/A & N/A & N/A & Harris (1957) \\
\hline
make use (of) & use & make & have & 6.55 & Dras, Dale (1995) \\
\hline
\end{tabular}
\caption{Support verb candidates chosen by the system} \label{tab3}
\end{table*}
}

\vspace{0.15in}

The test set and results are summarised 
in Table~\ref{tab3}; the table contains:

\begin{itemize}
\item the source text;
\item the corresponding verb, which the source can be rewritten as;
\item the reference from which the source text was taken;
\item the system's first choice candidate for support verb C1 for the source 
text's constituent nominalisation (i.e. the verb category with the highest
expected number of light verbs, SV''(j));
\item the system's second choice C2; and
\item the ratio of the expected number of light verbs for the first and 
second choices.
\end{itemize}

A second system was implemented based on the basic model of section 3.1; for comparison, Table 3 gives the results of this system.

{\small
\begin{table*}
\begin{tabular} {|l|l|c|l|l|l|} 
\hline
Source Text & Verb & SV'(j)\\
\hline
\hline
make an attempt & attempt & make \\
\hline
make a change & change & undergo \\
\hline
make a concession & concede & make\\
\hline
make a demand & demand & meet \\
\hline
make a distinction & distinguish & make \\
\hline
have a drink (of) & drink & become \\
\hline
have an effect (on) & affect & have \\
\hline
have a feeling & feel & express \\
\hline
make a gift (of) & give & have \\
\hline
do harm (to) & harm & cause \\
\hline
make a judgment & judge & make \\
\hline
have a knowledge (of) & know & have \\
\hline
make progress & progress & make \\
\hline
make a proposal & propose & reject \\
\hline
bear a resemblance (to) & resemble & bear \\
\hline
give a shove (to) & shove & N/A \\
\hline
have a snooze & snooze & N/A \\
\hline
make use (of) & use & make \\
\hline
\end{tabular}
\caption{Support verb candidates chosen under the basic model} \label{tab4}
\end{table*}
}

\subsection{Discussion}

\subsubsection{Analysis of results of the global information model}

Of the 18 examples, 13 choices of support verb match the corresponding 
one from the source text.  Of the five cases where the chosen SV did not match 
the source verb, one was actually valid: \lingform{harm} had \lingform{cause} 
as the proposed alternative.  This is an equally plausible support verb,
and in any case, \lingform{do} was the second choice by only a small
margin.  This is true for a number of cases: where there is an alternative 
support verb to the one used in the source text, the second alternative 
represents another plausible choice, and the frequency ratio margin 
is small (for example, for \lingform{change} and \lingform{resemblance}).

In three cases lack of data is a problem, resulting in the three N/A values
in Table~\ref{tab3}: there are no occurrences of  \lingform{snooze} 
or \lingform{shove} as direct objects of verbs in Grolier's, most 
probably because they belong to a more informal register than that 
used in encyclopedias.  Similarly, \lingform{have a drink} is an informal 
phrase that would not normally be found in an encyclopedia, as evidenced by 
there being only one occurrence of a governing verb for \lingform{drink}.

The system's worst performance was with the nominalisation \lingform{gift}.  
This appears to have occurred as \lingform{gift} is frequently 
concretised, as in \lingform{She has a great gift which has astounded her 
teachers} or \lingform{This deal includes a free gift!}  However,
only one of the 18 cases appears to be affected in this way.

\subsubsection{Comparison}

So, allowing alternative SVs,
and disregarding the cases where the genres of the test data differed from the
genre of the training (encyclopedia) data, the success rate is 14 of 15 (93\%),
using a 66Mb corpus.  By comparison, Grefenstette and Teufel~(1995) 
achieve plausible SVs for 7 of 10 cases (70\%), using a 134Mb corpus; and using
the basic model described in section 3 achieves plausible support verbs for 10 of
15 cases.  The higher results achieved by the method proposed in this paper
are statistically significant at the 10\% and 5\% levels respectively.  Stronger results
may be obtained given more test data---it is difficult to do better than an
improvement of 5\% significance with only 15 cases.  Developing a larger set
will require further work, as there is often disagreement about the validity of 
equivalence between SV constructions and full verbs.  The data do suggest, however, 
that this is worthwhile, particularly when it is noted that the higher success rate was achieved with a smaller corpus.

In general, the method seems to cover well both productive and idiomatic
SV constructions.  For example,
\lingform{have} is productive in light verb constructions, and the high global 
frequency will give a relatively high expected lightness rate; however, it 
does not eliminate the possibility of a low-frequency verb (like 
\lingform{bear}) being a support verb in cases where the SV construction is 
strongly idiomatic (as in \lingform{bear a resemblance}), where the low 
frequency in the standard population is counterbalanced by a high frequency in 
the target population.


\section{Conclusion}

In the process of calculating expected SVs,
a number of significant assumptions were made: 

\begin{itemize}
\item that concretised nominals would 
not have a significant impact compared with the effect of the standardisation; 
\item that in initially constructing the global 
information, all verbs can be taken as light; and
\item that the productivity of SVs allows the construction of a reasonable
standard population.
\end{itemize}

Notwithstanding these considerations, the experimental results demonstrate
that approximating the conditional probabilities, by the unconditional
probabilities derived from a large number of nominalisations, 
provides accurate choices for support verbs for individual nominalisations in the
test set.  The accuracy of the method appears to be superior to existing statistical 
methods which use only local information; and the method also involves only
minimal development effort, unlike existing semantic methods.

It is apparent that what
constitutes a valid light verb construction depends 
on the genre and register of the text.  Given that the test set was taken from
a wide range of sources, more accurate results for this test set could 
no doubt be obtained by using a corpus that was more representative 
of general English.
Also, more accurate results might be gained after further iterations of this 
process: once the most likely support verb in a given local information is 
determined, the global information can be regenerated using these, rather than 
the assumption of universal lightness of verbs.
Further work will look at 
implementing this iterative process, developing a larger set of test data for
evaluation purposes, and extending this method to 
other light 
constructions---light verbs with adjectival complements, and light nouns 
with post-modifiers.

Also, in order to successfully carry out a process of disambiguation on a
random text, the coverage of nominalisations and SVs needs to be greater;
a key aspect of future work is expanding the set of data to achieve this
better coverage.

\end{document}